\documentclass[10pt,twocolumn,letterpaper]{article}

\PassOptionsToPackage{table}{xcolor}

\usepackage{wacv}

\usepackage{makecell}
\usepackage{multirow}
\usepackage{multicol}
\usepackage{booktabs}
\usepackage{amsmath,amssymb,amsfonts}

\usepackage{tabularx}
\usepackage{pifont}
\newcolumntype{Y}{>{\centering\arraybackslash}X}
\definecolor{oursbg}{RGB}{235,237,246}
\newcommand{\ours}{\rowcolor{oursbg}}
\newcommand{\cmark}{{\color{ForestGreen!80}\ding{51}}}
\newcommand{\xmark}{{\color{BrickRed!80}\ding{55}}}
\newcommand{\pmbox}[1]{\raisebox{-.35ex}{\fontsize{5.5pt}{6.5pt}\selectfont$\pm$#1}}
\newcommand{\val}[2]{#1\pmbox{#2}}
\newcommand{\vb}[2]{\textbf{#1}\pmbox{#2}}
\newcommand{\vu}[2]{\underline{#1}\pmbox{#2}}

\definecolor{myRed}{RGB}{195,10,10}
\definecolor{myGreen}{RGB}{55,149,73}
\definecolor{myBlue}{RGB}{5,5,60}

\definecolor{Gray}{gray}{0.9}

\makeatletter
\renewcommand\paragraph{\@startsection{paragraph}{4}{\z@}{1.4ex \@plus 0.4ex \@minus 0.2ex}{-0.6em}{\normalfont\normalsize\bfseries}}
\makeatother

\definecolor{wacvblue}{rgb}{0.21,0.49,0.74}
\usepackage[pagebackref,breaklinks,colorlinks,allcolors=wacvblue]{hyperref}

\def\wacvPaperID{1422}
\def\confName{WACV}
\def\confYear{2027}

\title{A Multi-Agent Framework for Automated Machine Learning with Evidence-based Planning for Medical Image Analysis}

\author{Yiqing Shen, Mathias Unberath\\
Department of Computer Science, Johns Hopkins University\\
{\tt\small \{yshen92, unberath\}@jhu.edu}}

\begin{document}
\maketitle
\begin{abstract}

Deep learning has become the default tool for medical image analysis, driven by the open-source movement.
However, achieving the best performance on a given task remains difficult, with architecture selection and hyperparameter optimization (HPO) still considered ``more of an art than a science''.
Automated machine learning (AutoML) automates parts of model training, yet helps little in medical imaging: general-purpose approaches lack medical architectures and preprocessing, and medical-focused ones demand engineering expertise and remain confined to narrow tasks such as 3D segmentation.
We introduce MedicalMLOmni, a multi-agent AutoML framework that automates model training for diverse single- and multi-modal medical tasks through a natural language interface.
Specifically, MedicalMLOmni performs evidence-based planning by retrieving published results to guide model selection and configuration, and employs parallel HPO where multiple agents explore adjustment strategies concurrently.
MedicalMLOmni achieves the best score on 19 of 20 benchmarks; PMC-VQA is the exception, where a vision-language model with broader pre-training leads.
Backbone-matched controls attribute the gains to the framework, not the backbone.
By removing coding requirements, MedicalMLOmni lowers the barrier for domain experts with rich datasets but limited engineering expertise.

\end{abstract}

\section{Introduction}

Medical image analysis relies heavily on deep learning (DL).
However, deploying DL models for new clinical sites typically requires retraining with architecture modifications, custom preprocessing, and hyperparameter optimization (HPO) due to anatomical variability and protocol differences between institutions~\cite{su2024navigating}.
This process demands considerable time and expertise, limiting clinical adoption~\cite{thirunavukarasu2023comparative}.
Automated machine learning (AutoML) was therefore introduced to eliminate manual effort by automatically training DL models for new tasks.
General-purpose AutoML methods such as AutoGluon~\cite{autogluon} and AutoKeras~\cite{autokeras} lack medical-specific preprocessing and model architectures.
Medical-focused AutoML methods like Auto3DSeg partially address these gaps by configuring preprocessing and selecting medical-specific architectures based on dataset properties~\cite{monai}.
However, these approaches remain limited to specific scopes such as 3D segmentation, excluding other tasks like multi-modal fusion.
Most AutoML methods also require coding knowledge to operate, creating barriers for clinical users~\cite{autom3l}.

Large language models (LLMs) address these limitations through multi-agent systems that generate executable ML pipelines from natural language task descriptions without manual coding.
For example, AutoM3L~\cite{autom3l} employs LLMs to identify data and task types and generate corresponding DL training pipelines.
MLZero~\cite{mlzero} extends this approach with memory to reuse past experiences in similar problems.
However, these general-purpose AutoML agents lack medical knowledge and cannot select appropriate DL architectures or configurations for clinical applications.
Existing AutoML agents also configure hyperparameters through optimization methods such as Bayesian optimization~\cite{autogluon,autom3l}, without drawing on domain-specific knowledge from the literature.

We introduce MedicalMLOmni, a multi-agent AutoML framework for medical applications.
Multi-agent automation of medical model development is itself an active area~\cite{m3builder,maistro,autohealth,agenticmed,amid}.
What distinguishes MedicalMLOmni is therefore the source of its configuration decisions, as Tab.~\ref{tab:positioning} summarizes.
MedicalMLOmni supports diverse uni-/multi-modal tasks such as visual question answering (VQA) and segmentation through an extensible architecture that adapts to new task formulations.
MedicalMLOmni operates in two stages, where the planning stage analyzes the data and the user's query, while the execution stage implements the generated workflow.
Specifically, MedicalMLOmni performs evidence-based planning by searching the medical literature online, grounding model selection and hyperparameter configuration in published results rather than templates, presets, or the LLM's own prior.
For the execution stage, MedicalMLOmni employs parallel reasoning for efficient HPO, where each agent independently explores hyperparameter adjustment strategies from the literature before aggregation.

The major contributions are threefold.
First, we introduce a medical-specific multi-agent AutoML framework that automates DL model training for diverse tasks through natural language interfaces.
Second, we propose evidence-based planning that retrieves information from literature to inform architecture and configuration decisions.
Finally, we introduce parallel reasoning for more efficient HPO where multiple agents independently explore adjustment strategies before aggregation.
To isolate what the framework contributes, we further report a backbone-matched control, a domain-expert baseline, a retrieval-reproducibility study, and a contamination-controlled protocol that excludes benchmark-specific sources from retrieval.

\section{Related Work}

\paragraph{Agentic AutoML.}

General-purpose AutoML frameworks such as AutoGluon~\cite{autogluon} and AutoKeras~\cite{autokeras} search over model families and hyperparameters through Bayesian or evolutionary optimization.
AutoM3L~\cite{autom3l} and MLZero~\cite{mlzero} replace that search with an LLM that reads a task description and emits a training pipeline.
These systems are domain-agnostic, so the architectures, preprocessing, and loss functions they reach are those of the general vision and tabular literature.

\paragraph{Multi-Agent Systems for Medical ML.}

Several concurrent systems bring multi-agent automation to medical data.
M$^3$Builder~\cite{m3builder} coordinates four role-playing agents over a structured medical imaging workspace, reporting model-building success rates on radiology datasets.
mAIstro~\cite{maistro} orchestrates eight task agents for radiomics and deep learning through a natural language interface.
AutoHealth~\cite{autohealth} adds uncertainty estimation to autonomous health data modeling, and further frameworks target end-to-end clinical inference~\cite{agenticmed} and auditable model development~\cite{amid}.
MedicalMLOmni is therefore not the first multi-agent AutoML system for medical imaging, and Tab.~\ref{tab:positioning} states what distinguishes it from these systems.
Prior systems draw architectures, losses, and hyperparameters from templates, workspace resources, or LLM priors, whereas MedicalMLOmni grounds them in retrieved literature and reuses the same retrieval channel to drive hyperparameter refinement.

\paragraph{Retrieval-Augmented Configuration.}

Retrieval augmentation is established for question answering, and LLM agents for code generation already plan, simulate, and debug their own programs~\cite{islam2025codesim,shi2024code,adnan2025large}.
To our knowledge, applying retrieval to the configuration decisions of a training pipeline, and treating benchmark scores from the retrieved sources as the stopping criterion $\tau$, is not addressed by the systems above.

\begin{table}[t!]
\caption{\textbf{Concurrent multi-agent medical AutoML systems differ in where the training configuration originates.}
All of them share the agentic decomposition.
Config.\ source: where the architecture, loss and hyperparameters come from; Refinement: how a configuration is revised after a run.}
\label{tab:positioning}
\centering
\fontsize{8pt}{9.6pt}\selectfont
\setlength{\tabcolsep}{3pt}
\begin{tabularx}{\columnwidth}{@{}lYY@{}}
\toprule
System & Config.\ source & Refinement \\
\midrule
AutoM3L~\cite{autom3l} & LLM prior & Bayesian opt. \\
MLZero~\cite{mlzero} & LLM prior + memory & Episodic feedback \\
M$^3$Builder~\cite{m3builder} & Prompt templates & Exec.\ debugging \\
mAIstro~\cite{maistro} & Expert prompts & Exec.\ feedback \\
AutoHealth~\cite{autohealth} & LLM prior & Pred.\ uncertainty \\
AMID~\cite{amid} & LLM prior & Audit trail \\
\midrule
\ours MedicalMLOmni (Ours) & Literature retrieval & Literature retrieval \\
\bottomrule
\end{tabularx}
\end{table}

\section{Methods}

\begin{figure*}[tb]
\centering
\includegraphics[width=\linewidth]{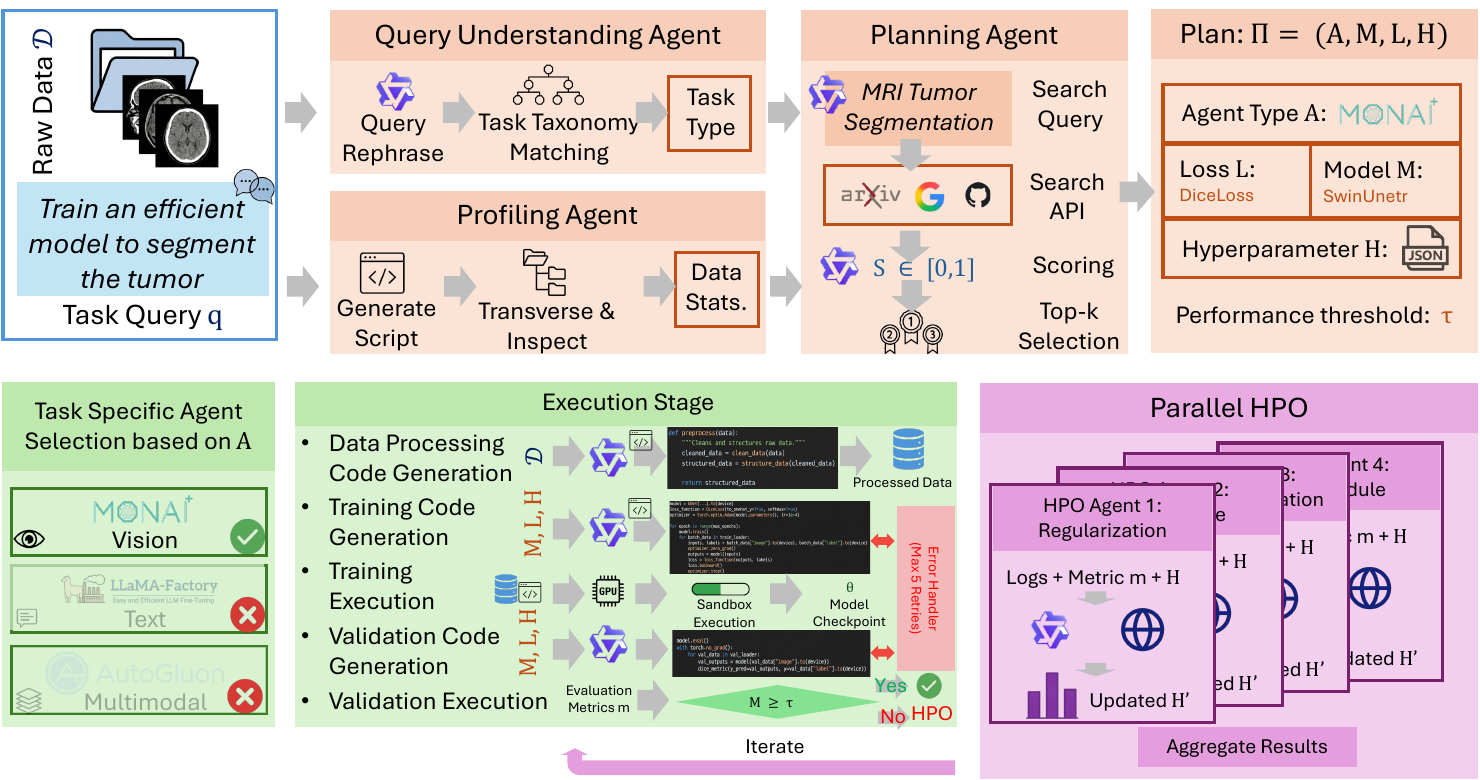}
\caption{
Overview of MedicalMLOmni showing evidence-based planning through literature retrieval, task-specific execution agents, and parallel hyperparameter optimization.
}
\label{fig:method}
\end{figure*}

\paragraph{Overview.}
As shown in Fig.~\ref{fig:method}, MedicalMLOmni operates in two stages.
Given raw data $\mathcal{D}$ (specified as the path to the data) and task description $q$, the planning stage generates an evidence-based training plan $\pi = (\mathcal{A}, \mathcal{M}, \mathcal{L}, \mathcal{H})$.
Specifically, $\mathcal{A}$ identifies the task agent type that determines which implementation framework to use based on task type and data modality, $\mathcal{M}$ defines the model architecture, $\mathcal{L}$ selects the loss function, and $\mathcal{H}$ contains model hyperparameters.
The execution stage implements $\pi$ through task-specific agents that generate and execute training code, producing trained model $\theta^*$, predictions $\hat{y}$, and validation metrics $m$.
When $m$ fails to meet expected performance thresholds $\tau$ specified in the planning stage, the parallel HPO refines $\pi$ through simultaneous literature-based hyperparameter adjustments.

\paragraph{Evidence-Based Planning Stage.}
The planning stage constructs $\pi$ through three agents that analyze task requirements, profile data characteristics, and retrieve evidence from the literature.
The query understanding agent prompts the LLM to parse $q$ by matching against pre-defined task taxonomies.
In parallel, the profiling agent generates Python scripts that traverse $\mathcal{D}$, inspect headers to detect data modality, and load samples to compute dataset properties such as count, dimensions, and memory requirements.
Together, the identified task type and the detected data modality determine $\mathcal{A}$, which routes to the appropriate implementation agent during execution.
The planning agent then constructs search queries by prompting the LLM to combine the extracted task type and data properties, submitting them to the search APIs~\cite{chen2024mindsearch,searchexpert}.
Retrieved results undergo LLM-based extraction to identify model architectures, loss functions, and hyperparameters.
Each configuration receives an LLM-calculated relevance score $s \in [0,1]$ that quantifies the alignment between the reported characteristics from the search results and the current data profile.
Instead of selecting a single source, the planning agent retrieves top-$k$ highest-scoring configurations to enable selective combination.
The planning agent then prompts the LLM to synthesize $\pi$ by selecting $\mathcal{M}$ and $\mathcal{L}$ from the highest-scored sources, while aggregating $\mathcal{H}$ across multiple sources to incorporate complementary recommendations.
The agent also extracts performance thresholds $\tau$ by prompting the LLM to identify validation metrics and benchmark scores reported for similar tasks and datasets in the retrieved results.
When no retrieved source reports a comparable benchmark, or when every source scores below $s{=}0.3$, $\tau$ falls back to a fixed-budget schedule that triggers exactly one HPO round.
A failed retrieval therefore degrades the framework to a single blind refinement, not an unbounded loop.
Because $\tau$ controls only when HPO fires and never enters the training objective, an over- or under-estimated $\tau$ trades compute against accuracy and does not bias the final model.
Tab.~\ref{tab:retrieval} quantifies this sensitivity by perturbing $\tau$ by $\pm$10\%.
MedicalMLOmni can optionally present $\pi$ with supporting evidence sources for user confirmation before execution.

\paragraph{Execution Stage.}
The execution stage routes to task-specific agents based on $\mathcal{A}$ identified during planning.
Specifically, MedicalMLOmni instantiates vision task agents via \texttt{MONAI}~\cite{monai}, text task agents via \texttt{LLaMA-Factory}~\cite{llamafactory}, and multimodal task agents via \texttt{AutoMM}~\cite{autogluon}, and further agents can be registered for new task formulations.
The Qwen3-30B backbone acts only as the planning and code-generating agent; it never contributes weights to the trained model.
For VQA, the framework fine-tunes whichever vision-language model the plan selects through \texttt{LLaMA-Factory}, sharing no weights or pre-training with the Qwen3-VL baseline beyond both being publicly released checkpoints.
The output of each run is therefore an ordinary task model that can be deployed without the agent, and Tab.~\ref{tab:backbone} reports what the same backbone achieves once the framework is removed.
Each task agent implements a five-step pipeline in which an LLM generates Python code conditioned on $\pi$ and the outputs from the preceding steps~\cite{islam2025codesim}.
The pipeline begins with the generation of data processing code, where the LLM generates scripts that load $\mathcal{D}$, apply transformations, and create train/validation/test splits if they are not available.
These scripts are executed to produce preprocessed data paths, which then condition the LLM to generate training code that instantiates $\mathcal{M}$ with $\mathcal{L}$ and $\mathcal{H}$ from $\pi$.
The training code runs in a sandbox environment with resource monitoring, producing checkpoint $\theta^*$ and execution logs.
The sandbox is network-isolated and mounts $\mathcal{D}$ read-only, so patient data never leaves the local execution environment.
Retrieval runs on a separate channel that transmits only the derived data profile (modality, voxel counts, class counts, and intensity statistics), never image content, filenames, or identifiers, keeping the online component free of protected health information.
Subsequently, the checkpoint path conditions the LLM to generate validation scripts that load $\theta^*$ and compute metrics specified in $\pi$.
Executing these validation scripts produces validation metrics $m$ and predictions $\hat{y}$, which are compared against thresholds $\tau$ to determine whether HPO is needed.
When code execution fails at any step (\textit{e}.\textit{g}., out-of-memory errors, dimension mismatches), the LLM analyzes error messages and execution logs to generate corrected code~\cite{shi2024code,adnan2025large}.
The LLM retrieves debugging strategies from online resources and incorporates these fixes into regenerated code.
This debugging process iterates with a maximum of five retry attempts before escalating to user review.

\paragraph{Parallel Hyperparameter Optimization.}
When validation metrics $m$ fail to meet performance thresholds $\tau$ extracted during planning, MedicalMLOmni initiates parallel HPO to refine the configuration.
Formally, MedicalMLOmni spawns $n$ HPO agents that independently explore hyperparameter modification strategies, each receiving the current plan $\pi$, validation metrics $m$, and execution logs as input.
Each HPO agent constructs literature search queries by prompting the LLM to combine underperforming metric names, current hyperparameter values from $\mathcal{H}$, and task properties, submitting them to the search APIs~\cite{chen2024mindsearch,searchexpert} used during planning.
HPO agents are prompted to focus on different hyperparameter categories such as optimization adjustments, regularization strategies, or architecture-specific parameters to ensure diverse exploration.
Retrieved results undergo extraction to identify adjustment strategies reported for similar tasks, which each agent scores based on relevance to the observed performance gap.
Each agent generates the candidate configuration $\mathcal{H}'_i$ along with confidence $C_i \in [0,1]$ reflecting cross-source consistency and task alignment, and uncertainty $U_i \in [0,1]$ quantifying search quality and hyperparameter variance across sources.
MedicalMLOmni aggregates candidates $\{\mathcal{H}'_1, \mathcal{H}'_2, \ldots, \mathcal{H}'_n\}$ with weights $w_i = C_i / (1 + U_i)$, prompting the LLM to synthesize refined configuration $\mathcal{H}^*$ by selecting consensus hyperparameters, resolving conflicts through weighted voting, and combining complementary adjustments.
This updates the plan to $\pi' = (\mathcal{A}, \mathcal{M}, \mathcal{L}, \mathcal{H}^*)$, which re-enters execution for retraining.
For high-uncertainty scenarios where $\max_i(U_i) > 0.7$, MedicalMLOmni can optionally present candidate configurations with their confidence and uncertainty scores to users for validation before retraining.
The parallel HPO process iterates until $m$ satisfies $\tau$ or the maximum iterations are reached.

\section{Experiments}

\paragraph{Datasets and Evaluation Metrics.}
\label{sec:setup}
We evaluate MedicalMLOmni across three categories of medical imaging tasks.
For segmentation, we assess performance on eight benchmarks that span various anatomical regions and imaging modalities:
AMOS~\cite{amos} for multi-organ segmentation from CT and MRI; BraTS23~\cite{braints} and ACDC~\cite{acdc} for brain tumor and cardiac segmentation in MRI; KiTS23~\cite{kits21}, BTCV~\cite{btcv}, TotalSegmentator~\cite{totalsegmentator}, and Touchstone~\cite{touchstone} for kidney, abdominal organ, anatomical structures, and abdominal segmentation in CT; and MedSeg~\cite{medsegbench} for ultrasound, MRI, and X-ray segmentation.
For visual question answering, we evaluate on seven benchmarks: VQA-RAD~\cite{vqarad} and GEMeX~\cite{gemex} for radiology X-ray, SLAKE~\cite{slake} and PMC-VQA~\cite{pmcvqa} for multilingual and multi-modal VQA; PathVQA~\cite{pathvqa} for pathology image analysis, MedXpertQA~\cite{medxpert} for expert-level clinical assessment; and Kvasir-VQA~\cite{kvasir} for gastrointestinal endoscopy analysis.
For multimodal fusion, we evaluate on five benchmarks: BraTS-Sur~\cite{brats}, MIMIC~\cite{mimic}, and NLST~\cite{nlst} combine medical imaging with clinical data, text reports, and electronic health records for survival, mortality, and lung cancer risk prediction; ADNI~\cite{adni} merges MRI and PET imaging for Alzheimer's diagnosis; ISIC~\cite{isic} fuses dermoscopic images with metadata for melanoma classification.
We report the Dice similarity coefficient for segmentation, accuracy for VQA, and task-specific metrics, namely accuracy, AUC, or pAUC (patient-level AUC) for multimodal benchmarks.
We use Qwen3-30B as the backbone.
The retrieval corpus is restricted to a frozen PubMed and arXiv snapshot (January 1, 2026) to ensure reproducibility and guard against contamination.

\begin{table*}[tb]
\caption{\textbf{MedicalMLOmni reaches the highest Dice on all eight segmentation benchmarks.}
Dice $\uparrow$ as mean $\pm$ standard deviation over repeated runs; the second header row gives the imaging modality.
Bold: best; underline: second best.}
\label{tab:vision_comparison}
\centering
\fontsize{8pt}{9.6pt}\selectfont
\setlength{\tabcolsep}{1.5pt}
\begin{tabularx}{\textwidth}{@{}lYYYYYYY>{\hsize=1.12\hsize}Y>{\hsize=0.88\hsize}Y@{}}
\toprule
Method & AMOS~\cite{amos} & BraTS23~\cite{braints} & KiTS23~\cite{kits21} & BTCV~\cite{btcv} & ACDC~\cite{acdc} & TotalSeg~\cite{totalsegmentator} & MedSeg~\cite{medsegbench} & Touchstone~\cite{touchstone} & Avg. \\
 & CT/MRI & MRI & CT & CT & MRI & CT & Mixed & CT & \\
\midrule
nnU-Net~\cite{nnunet} & \vu{0.893}{0.004} & \val{0.929}{0.003} & \val{0.831}{0.005} & \val{0.866}{0.004} & \val{0.914}{0.003} & \vu{0.921}{0.004} & \vu{0.884}{0.005} & \vu{0.889}{0.004} & \vu{0.891}{0.004} \\
MedNeXt-L~\cite{mednext} & \val{0.888}{0.005} & \vu{0.936}{0.003} & \val{0.825}{0.006} & \vu{0.875}{0.004} & \vu{0.920}{0.003} & \val{0.898}{0.005} & \val{0.876}{0.006} & \val{0.883}{0.005} & \val{0.888}{0.005} \\
STU-Net-L~\cite{stunet} & \val{0.891}{0.004} & \val{0.931}{0.004} & \val{0.820}{0.007} & \val{0.852}{0.006} & \val{0.895}{0.005} & \val{0.891}{0.005} & \val{0.863}{0.007} & \val{0.871}{0.006} & \val{0.877}{0.005} \\
SegResNet~\cite{myronenko20183d} & \val{0.861}{0.006} & \val{0.910}{0.005} & \val{0.808}{0.008} & \val{0.819}{0.007} & \val{0.889}{0.006} & \val{0.878}{0.006} & \val{0.857}{0.008} & \val{0.856}{0.007} & \val{0.860}{0.007} \\
VISTA3D~\cite{vista3d} & \val{0.850}{0.007} & \val{0.912}{0.005} & \vu{0.837}{0.006} & \val{0.845}{0.006} & \val{0.903}{0.005} & \val{0.912}{0.005} & \val{0.871}{0.006} & \val{0.862}{0.007} & \val{0.874}{0.006} \\
Mamba-UNet~\cite{mambaunet} & \val{0.870}{0.006} & \val{0.904}{0.006} & \val{0.813}{0.007} & \val{0.858}{0.006} & \val{0.908}{0.005} & \val{0.874}{0.007} & \val{0.861}{0.007} & \val{0.848}{0.008} & \val{0.867}{0.007} \\
SwinUNETR~\cite{swinunetr} & \val{0.839}{0.008} & \val{0.895}{0.007} & \val{0.798}{0.009} & \val{0.840}{0.007} & \val{0.882}{0.007} & \val{0.860}{0.008} & \val{0.849}{0.009} & \val{0.833}{0.008} & \val{0.850}{0.008} \\
AutoKeras~\cite{autokeras} & \val{0.783}{0.012} & \val{0.857}{0.010} & \val{0.758}{0.014} & \val{0.792}{0.011} & \val{0.839}{0.010} & \val{0.731}{0.015} & \val{0.803}{0.012} & \val{0.767}{0.013} & \val{0.791}{0.012} \\
AutoGluon~\cite{autogluon} & \val{0.771}{0.013} & \val{0.849}{0.011} & \val{0.751}{0.015} & \val{0.785}{0.012} & \val{0.832}{0.011} & \val{0.769}{0.014} & \val{0.794}{0.013} & \val{0.759}{0.014} & \val{0.789}{0.013} \\
MLZero~\cite{mlzero} & \val{0.774}{0.012} & \val{0.851}{0.010} & \val{0.750}{0.014} & \val{0.786}{0.011} & \val{0.830}{0.011} & \val{0.771}{0.013} & \val{0.796}{0.012} & \val{0.760}{0.013} & \val{0.790}{0.012} \\
AutoM3L~\cite{autom3l} & \val{0.765}{0.014} & \val{0.841}{0.012} & \val{0.745}{0.016} & \val{0.778}{0.013} & \val{0.825}{0.012} & \val{0.762}{0.015} & \val{0.787}{0.014} & \val{0.752}{0.015} & \val{0.782}{0.014} \\
Auto3DSeg~\cite{monai} & \val{0.877}{0.005} & \val{0.917}{0.004} & \val{0.832}{0.006} & \val{0.855}{0.005} & \val{0.903}{0.005} & \val{0.904}{0.005} & \val{0.871}{0.006} & \val{0.874}{0.006} & \val{0.879}{0.005} \\
\midrule
\ours MedicalMLOmni (Ours) & \vb{0.908}{0.003} & \vb{0.951}{0.002} & \vb{0.852}{0.004} & \vb{0.890}{0.003} & \vb{0.935}{0.002} & \vb{0.939}{0.003} & \vb{0.899}{0.004} & \vb{0.905}{0.003} & \vb{0.910}{0.003} \\
\bottomrule
\end{tabularx}
\end{table*}

\begin{figure*}[tb]
\centering
\includegraphics[width=\linewidth]{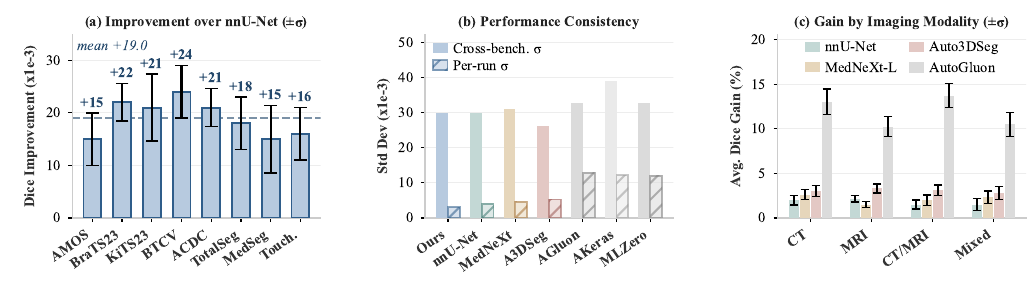}
\caption{
Segmentation analysis.
(a)~Per-benchmark Dice improvement over nnU-Net with propagated standard deviation as error bars.
(b)~Cross-benchmark standard deviation (solid) and average per-run standard deviation (hatched) for each method, measuring performance consistency.
(c)~Average Dice gain over four baselines grouped by imaging modality, with error bars indicating propagated uncertainty.
}
\label{fig:seg_analysis}
\end{figure*}

\paragraph{Medical Image Segmentation Results.}
Tab.~\ref{tab:vision_comparison} presents the segmentation performance across eight benchmarks.
MedicalMLOmni achieves the highest Dice scores on all datasets, with an average of 0.910, outperforming the second-best method nnU-Net~\cite{nnunet} by 0.019.
It also exceeds MedNeXt-L~\cite{mednext} (0.888), STU-Net-L~\cite{stunet} (0.877), and VISTA3D~\cite{vista3d} (0.874) without manual architecture engineering, exceeds the medical-specific Auto3DSeg~\cite{monai} by 0.031, and improves on general-purpose AutoML by 0.119 to 0.128.
The specialized baselines follow their published recipes, not a budget matched to ours, so Tab.~\ref{tab:backbone} repeats the comparison with the tuning budget equalized.

Fig.~\ref{fig:seg_analysis} analyzes these scores per benchmark, per method, and per modality.
Fig.~\ref{fig:seg_analysis}(a) shows per-benchmark gains over nnU-Net~\cite{nnunet}, largest on BTCV (+24) and BraTS23 (+22) and averaging +19.0 ($\times10^{-3}$).
Fig.~\ref{fig:seg_analysis}(b) shows the lowest per-run standard deviation of all methods, about a quarter of that of general-purpose AutoML, while the spread across benchmarks is comparable for all methods.
Fig.~\ref{fig:seg_analysis}(c) decomposes the gains by modality, which are positive on every modality against all four baselines.

\begin{figure*}[tb]
\centering
\includegraphics[width=\linewidth]{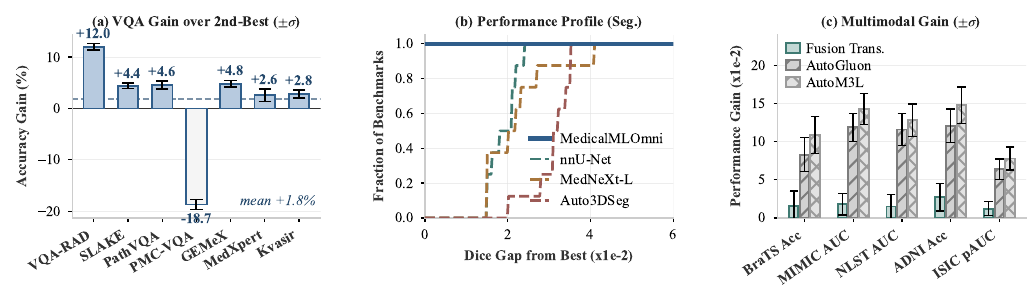}
\caption{
VQA and multimodal analysis.
(a)~Per-benchmark VQA accuracy gain of MedicalMLOmni over the respective second-best method, with propagated standard deviation as error bars.
(b)~Performance profile showing the fraction of segmentation benchmarks on which each method is within a given Dice gap from the best.
(c)~Per-benchmark performance gain of MedicalMLOmni over three multimodal baselines, with propagated uncertainty.
}
\label{fig:vqa_mm_analysis}
\end{figure*}

\begin{table*}[tb]
\caption{\textbf{MedicalMLOmni is best on six of the seven VQA benchmarks and on average.}
Accuracy $\uparrow$ in \% as mean $\pm$ standard deviation over repeated runs; the second header row gives the image domain.
Bold: best; underline: second best.}
\label{tab:vqa_comparison}
\centering
\fontsize{8pt}{9.6pt}\selectfont
\setlength{\tabcolsep}{1.5pt}
\begin{tabularx}{\textwidth}{@{}lYYYYYYYY@{}}
\toprule
Method & VQA-RAD~\cite{vqarad} & SLAKE~\cite{slake} & PathVQA~\cite{pathvqa} & PMC-VQA~\cite{pmcvqa} & GEMeX~\cite{gemex} & MedXpert~\cite{medxpert} & Kvasir~\cite{kvasir} & Avg. \\
 & X-ray & Multi & Path & Multi & X-ray & Clinical & Endo & \\
\midrule
MedVInT-TD~\cite{pmcvqa} & \vu{73.9}{0.6} & \val{84.7}{0.6} & \val{55.8}{0.9} & \val{35.2}{0.9} & \val{70.8}{0.7} & \val{24.6}{1.0} & \val{61.3}{0.8} & \val{58.0}{0.8} \\
MedVInT-TE~\cite{pmcvqa} & \val{69.5}{0.7} & \vu{88.4}{0.5} & \val{48.8}{1.0} & \val{36.6}{0.9} & \val{68.2}{0.8} & \val{23.8}{1.1} & \val{48.8}{1.1} & \val{54.9}{0.9} \\
BiomedGPT~\cite{biomedgpt} & \val{61.1}{0.8} & \val{86.3}{0.5} & \val{52.3}{0.9} & \val{24.1}{1.1} & \val{66.5}{0.8} & \val{21.7}{1.2} & \val{58.6}{0.9} & \val{52.9}{0.9} \\
LLaVA-Med-v1.5~\cite{llavamed} & \val{64.7}{0.7} & \val{82.1}{0.6} & \val{55.5}{0.9} & \val{29.1}{1.0} & \val{73.4}{0.6} & \vu{28.6}{1.0} & \vu{68.7}{0.7} & \val{57.4}{0.8} \\
AutoGluon~\cite{autogluon} & \val{40.8}{1.3} & \val{59.5}{1.2} & \val{35.1}{1.5} & \val{15.8}{1.4} & \val{46.4}{1.3} & \val{14.2}{1.5} & \val{37.9}{1.4} & \val{35.7}{1.4} \\
AutoKeras~\cite{autokeras} & \val{39.5}{1.4} & \val{57.2}{1.3} & \val{37.4}{1.5} & \val{14.6}{1.5} & \val{45.1}{1.4} & \val{13.8}{1.6} & \val{36.2}{1.5} & \val{34.8}{1.5} \\
MLZero~\cite{mlzero} & \val{43.7}{1.2} & \val{61.0}{1.1} & \val{39.1}{1.4} & \val{17.4}{1.3} & \val{48.7}{1.2} & \val{17.1}{1.4} & \val{40.4}{1.3} & \val{38.2}{1.3} \\
Qwen2.5-VL~\cite{qwen2vl} & \val{68.4}{0.5} & \val{78.0}{0.4} & \val{66.0}{0.6} & \vu{55.8}{0.7} & \val{71.3}{0.7} & \val{25.7}{1.0} & \val{63.8}{0.8} & \val{61.3}{0.7} \\
Qwen3-VL~\cite{qwen3vl} & \val{71.8}{0.6} & \val{81.3}{0.5} & \vu{69.2}{0.7} & \vb{58.4}{0.8} & \vu{74.1}{0.6} & \val{28.3}{0.9} & \val{66.5}{0.7} & \vu{64.2}{0.7} \\
\midrule
\ours MedicalMLOmni (Ours) & \vb{85.9}{0.3} & \vb{92.8}{0.2} & \vb{73.8}{0.4} & \val{39.7}{0.6} & \vb{78.9}{0.3} & \vb{31.2}{0.7} & \vb{71.5}{0.4} & \vb{67.7}{0.4} \\
\bottomrule
\end{tabularx}
\end{table*}

\paragraph{Medical Visual Question Answering Results.}
Beyond segmentation, we further evaluate question answering across seven benchmarks spanning radiology, pathology, endoscopy, and clinical domains, as shown in Tab.~\ref{tab:vqa_comparison}.
MedicalMLOmni reaches 67.7\% average accuracy against 64.2\% for fine-tuned Qwen3-VL~\cite{qwen3vl}, 58.0\% for MedVInT-TD~\cite{pmcvqa}, 57.4\% for LLaVA-Med-v1.5~\cite{llavamed}, and 52.9\% for BiomedGPT~\cite{biomedgpt}, and improves accuracy over general-purpose AutoML by more than 29\%.
It is best on six of the seven benchmarks; PMC-VQA is the exception and is analyzed below.

Fig.~\ref{fig:vqa_mm_analysis} analyzes these results per benchmark.
Fig.~\ref{fig:vqa_mm_analysis}(a) shows the per-benchmark gain over the respective second-best method, largest on VQA-RAD (+12.0\%) and GEMeX (+4.8\%), with PMC-VQA the single negative case ($-$18.7\%).
Fig.~\ref{fig:vqa_mm_analysis}(b) plots a performance profile over the segmentation benchmarks, where MedicalMLOmni holds every best score at zero tolerance while nnU-Net~\cite{nnunet} needs a gap of 2.4$\times10^{-2}$.
Fig.~\ref{fig:vqa_mm_analysis}(c) shows the multimodal gains, which are largest on ADNI and MIMIC and much larger over general-purpose AutoML than over the specialized Fusion Transformer~\cite{gnncnn}.

\begin{table*}[tb]
\caption{\textbf{MedicalMLOmni ranks first on all five multimodal benchmarks.}
Primary metrics vary by task and are given as mean $\pm$ standard deviation over repeated runs; average rank is computed per benchmark and then averaged.
Bold: best; underline: second best.}
\label{tab:multimodal_comparison}
\centering
\fontsize{8pt}{9.6pt}\selectfont
\setlength{\tabcolsep}{2pt}
\begin{tabularx}{0.82\textwidth}{@{}lYYYYYc@{}}
\toprule
Method & BraTS-Sur~\cite{brats} & MIMIC~\cite{mimic} & NLST~\cite{nlst} & ADNI~\cite{adni} & ISIC~\cite{isic} & Avg. \\
 & MRI+Clin. & X-ray+Text & CT+EHR & MRI+PET & Img+Meta & rank \\
 & Acc $\uparrow$ & AUC $\uparrow$ & AUC $\uparrow$ & Acc $\uparrow$ & pAUC $\uparrow$ & $\downarrow$ \\
\midrule
Fusion Trans.~\cite{gnncnn} & \vu{0.712}{0.015} & \vu{0.853}{0.011} & \vu{0.847}{0.012} & \vu{0.812}{0.014} & \vu{0.197}{0.007} & \underline{2.00} \\
AutoGluon~\cite{autogluon} & \val{0.645}{0.019} & \val{0.752}{0.016} & \val{0.746}{0.018} & \val{0.718}{0.019} & \val{0.145}{0.012} & 3.00 \\
AutoKeras~\cite{autokeras} & \val{0.632}{0.020} & \val{0.741}{0.017} & \val{0.732}{0.019} & \val{0.705}{0.020} & \val{0.138}{0.013} & 4.40 \\
MLZero~\cite{mlzero} & \val{0.627}{0.020} & \val{0.735}{0.017} & \val{0.741}{0.018} & \val{0.698}{0.020} & \val{0.134}{0.013} & 4.80 \\
AutoM3L~\cite{autom3l} & \val{0.619}{0.021} & \val{0.728}{0.018} & \val{0.734}{0.019} & \val{0.691}{0.021} & \val{0.131}{0.014} & 5.80 \\
\midrule
\ours MedicalMLOmni (Ours) & \vb{0.728}{0.012} & \vb{0.871}{0.009} & \vb{0.862}{0.010} & \vb{0.839}{0.011} & \vb{0.209}{0.006} & \textbf{1.00} \\
\bottomrule
\end{tabularx}
\end{table*}

\paragraph{Multimodal Task Results.}
Tab.~\ref{tab:multimodal_comparison} presents performance on five benchmarks that require the integration of multiple modalities.
Since these benchmarks employ different evaluation metrics that cannot be directly averaged, we report average rank, computed per benchmark and averaged across all five tasks.
MedicalMLOmni achieves an average rank of 1.00, consistently ranking first across all benchmarks and outperforming the second-best method Fusion Transformer~\cite{gnncnn} with an average rank of 2.00.
General-purpose AutoML methods rank much lower, from AutoGluon~\cite{autogluon} at 3.00 to AutoM3L~\cite{autom3l} at 5.80.
Unlike task-specific fusion architectures that require manual design per modality combination, MedicalMLOmni identifies the integration strategy automatically.

\paragraph{Computational Cost.}

MedicalMLOmni reaches the highest Dice at 18 GPU-hours, against 42 for nnU-Net~\cite{nnunet} and 64 for STU-Net-L~\cite{stunet}, while being the only method leading segmentation, VQA, and multimodal fusion simultaneously.
Tab.~\ref{tab:cost_breakdown} breaks the budget down by stage; Fig.~\ref{fig:cost} gives the full trade-off and cross-task analysis.

\begin{figure*}[tb]
\centering
\includegraphics[width=\linewidth]{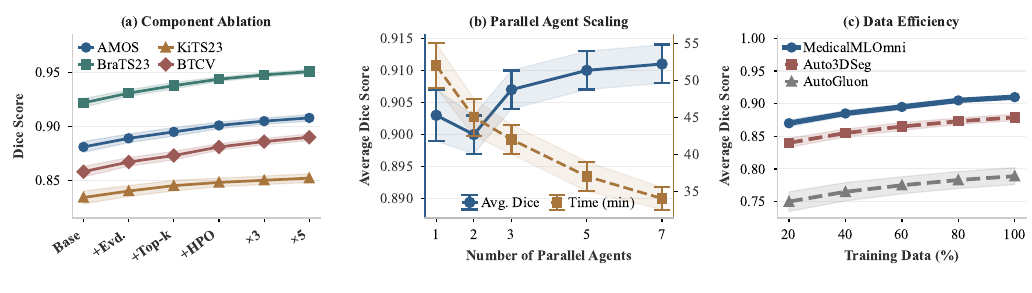}
\caption{
Ablation analysis on:
(a) Component ablation of Tab.~\ref{tab:ablation_segmentation} on four representative benchmarks, where $\times$3 and $\times$5 denote three and five parallel HPO agents.
(b) Parallel agent scaling effects on performance and time.
(c) Data efficiency across different training set sizes.
}
\label{fig:ablation}
\end{figure*}

\begin{table*}[tb]
\caption{\textbf{Every component adds Dice on every segmentation benchmark.}
Evd.: evidence-based retrieval; Top-$k$: multi-source synthesis; HPO: hyperparameter optimization; Agents: number of parallel HPO agents.
Dice $\uparrow$; bold: best; underline: second best.}
\label{tab:ablation_segmentation}
\centering
\fontsize{8pt}{9.6pt}\selectfont
\setlength{\tabcolsep}{2pt}
\begin{tabularx}{\textwidth}{@{}ccccYYYYYYYYY@{}}
\toprule
Evd. & Top-$k$ & HPO & Agents & AMOS & BraTS & KiTS & BTCV & ACDC & TSeg & MSeg & Touch & Avg. \\
\midrule
\xmark & \xmark & \xmark & 0 & .881 & .922 & .834 & .858 & .907 & .908 & .873 & .877 & .883 \\
\cmark & \xmark & \xmark & 0 & .889 & .931 & .840 & .867 & .915 & .915 & .881 & .885 & .890 \\
\cmark & \cmark & \xmark & 0 & .895 & .938 & .845 & .873 & .921 & .922 & .887 & .892 & .897 \\
\cmark & \cmark & \cmark & 1 & .901 & .944 & .848 & .881 & .927 & .929 & .893 & .898 & .903 \\
\cmark & \cmark & \cmark & 3 & \underline{.905} & \underline{.948} & \underline{.850} & \underline{.886} & \underline{.931} & \underline{.934} & \underline{.896} & \underline{.902} & \underline{.907} \\
\midrule
\ours \cmark & \cmark & \cmark & 5 & \textbf{.908} & \textbf{.951} & \textbf{.852} & \textbf{.890} & \textbf{.935} & \textbf{.939} & \textbf{.899} & \textbf{.905} & \textbf{.910} \\
\bottomrule
\end{tabularx}
\end{table*}

\paragraph{Ablation Study.}
Tab.~\ref{tab:ablation_segmentation} shows the contribution of each component of MedicalMLOmni across eight segmentation benchmarks.
Without evidence-based retrieval the average Dice is 0.883, already above general-purpose AutoML, so the task-specific agent architecture contributes independently of literature-based planning.
Retrieval adds 0.007, top-$k$ synthesis a further 0.007, sequential HPO reaches 0.903, and parallel HPO reaches 0.907 with three agents and 0.910 with five, showing consistent but diminishing returns.
Each added component improves all eight benchmarks.
Planning and retrieval together account for about half the total gain (0.014 of 0.027 Dice), with parallel HPO contributing the remainder, so neither stage dominates.

Fig.~\ref{fig:ablation} provides further analysis of the HPO dynamics and data efficiency.
Fig.~\ref{fig:ablation}(a) traces these configurations on four representative benchmarks, where each added component raises Dice.
Fig.~\ref{fig:ablation}(b) shows optimization time falling from 52 to 34 minutes between sequential refinement and seven agents, a 1.53$\times$ speedup whose sublinearity reflects aggregation overhead.
Fig.~\ref{fig:ablation}(c) varies the training set from 20\% to 100\%, where MedicalMLOmni leads Auto3DSeg~\cite{monai} and AutoGluon~\cite{autogluon} in every regime.

\begin{table}[t!]
\caption{\textbf{The framework, not the backbone, accounts for the improvement.}
The upper block removes MedicalMLOmni with the LLM fixed; the lower block swaps the LLM with MedicalMLOmni fixed, and the gains persist outside the Qwen3 family.
Dice: segmentation; VQA: accuracy in \%; MM: multimodal average rank; $^\dagger$tuning budget matched to our 18 GPU-hours.
Bold: best; underline: second best.}
\label{tab:backbone}
\centering
\fontsize{8pt}{9.6pt}\selectfont
\setlength{\tabcolsep}{2pt}
\begin{tabularx}{\columnwidth}{@{}llYYY@{}}
\toprule
Setting & Backbone & Dice $\uparrow$ & VQA $\uparrow$ & MM $\downarrow$ \\
\midrule
Default (no framework) & Qwen3-30B & 0.795 & 59.3 & 3.80 \\
Planning w/o evidence & Qwen3-30B & 0.883 & 63.4 & 2.20 \\
Best baseline$^\dagger$ & Task-specific & 0.898 & 65.1 & 1.40 \\
Domain expert & Task-specific & 0.901 & 66.5 & 1.20 \\
\midrule
MedicalMLOmni & Qwen3-8B & 0.885 & 62.1 & 2.60 \\
MedicalMLOmni & Llama-3.3-70B & 0.907 & \underline{68.2} & 1.10 \\
MedicalMLOmni & GPT-4o & \textbf{0.914} & \textbf{69.5} & \textbf{1.00} \\
\midrule
\ours MedicalMLOmni (Ours) & Qwen3-30B & \underline{0.910} & 67.7 & \textbf{1.00} \\
\bottomrule
\end{tabularx}
\end{table}

\paragraph{Backbone Control and Human Baseline.}

Tab.~\ref{tab:backbone} isolates the framework from the backbone by holding one factor fixed at a time.
Leaving Qwen3-30B to configure training from its prior yields 0.795 Dice, 59.3\% accuracy, and rank 3.80, so MedicalMLOmni contributes 0.115 Dice and 8.4\% accuracy over its own backbone, exceeding any margin over a specialized baseline.
Restoring planning without retrieval recovers most of that gap at 0.883 Dice and 63.4\%, attributing the remaining 0.027 Dice and 4.3\% accuracy to evidence.
Granting the strongest baseline a budget matched to our 18 GPU-hours lifts it from 0.891 to 0.898 Dice, narrowing our margin to 0.012, so a minority of the gap reflects unequal effort.
Against a domain expert configuring every benchmark manually, it slightly exceeds manual practice (0.910 vs.\ 0.901 Dice, 67.7\% vs.\ 66.5\%), the most relevant comparison for a system designed to replace that effort.
Swapping the backbone while holding the framework fixed shows that the gains are not Qwen3-specific.
Llama-3.3-70B and GPT-4o reach 0.907 and 0.914 Dice (68.2\% and 69.5\% VQA), within 0.003 Dice of or above the Qwen3-30B configuration, and even Qwen3-8B retains 0.885 Dice, above every AutoML baseline in Tab.~\ref{tab:vision_comparison}.
The framework therefore transfers across backbones and improves with stronger ones, allowing users to choose a backbone by licensing, latency, or cost without revalidating the framework.

\begin{table}[b!]
\caption{\textbf{A frozen snapshot makes planning exactly reproducible.}
Most of the advantage survives when benchmark-specific sources are withheld.
Overlap: mean Jaccard index between the top-$k$ source sets of independent retrieval runs; Agreement: fraction of runs selecting the same architecture and loss; Dice: mean over the eight segmentation benchmarks.}
\label{tab:retrieval}
\centering
\fontsize{8pt}{9.6pt}\selectfont
\setlength{\tabcolsep}{3pt}
\begin{tabularx}{\columnwidth}{@{}llYYY@{}}
\toprule
Corpus & Variation & Overlap $\uparrow$ & Agreement $\uparrow$ & Dice $\uparrow$ \\
\midrule
Live API & Baseline & 0.62 & 84\% & 0.910 \\
Frozen & Baseline & 1.00 & 100\% & 0.910 \\
Frozen & Paraphrased & 0.88 & 96\% & 0.908 \\
\midrule
Frozen & Bench.\ excl. & 0.58 & 80\% & 0.904 \\
Frozen & Pre-bench. & 0.53 & 76\% & 0.897 \\
\midrule
Frozen & $\tau$ $\pm$10\% & 1.00 & 100\% & 0.895 \\
\bottomrule
\end{tabularx}
\end{table}

\paragraph{Retrieval Reproducibility and Contamination.}

Tab.~\ref{tab:retrieval} reports that evidence-based planning is reproducible once the corpus is pinned, and that its advantage survives when benchmark-specific knowledge is withheld.
A live API gives a source-set overlap of only 0.62 across independent runs, yet plans agree 84\% of the time, indicating that synthesis is more stable than retrieval.
Pinning a corpus snapshot makes both exact: 1.00 overlap, 100\% agreement, and 0.910 Dice.
Paraphrasing the queries costs only 0.002 Dice, so we adopt the snapshot, not a live endpoint, as the reproducible configuration.
Turning to contamination, excluding every source that reports results on the benchmark being solved costs 0.006 Dice, and restricting to pre-benchmark sources costs 0.013, leaving 0.897.
This most restrictive setting still exceeds published nnU-Net (0.891) yet no longer separates from the budget-matched 0.898 of Tab.~\ref{tab:backbone}, confirming that part of the advantage derives from literature about the benchmarks themselves.
Perturbing $\tau$ by $\pm$10\% costs 0.015 Dice, the largest single sensitivity we measure, because HPO then fires too early or too late while the training objective stays unchanged.
The same design keeps patient data local: training runs in a network-isolated sandbox, retrieval sends only the derived data profile, and a locally mirrored corpus can remove the outbound path entirely at the cost of corpus freshness.

Tab.~\ref{tab:whatretrieved} of the supplementary records what the planner retrieves per task family.
Segmentation plans converge on an nnU-Net-style 3D U-Net with a DiceCE loss and match the best published recipe on 88\% of the benchmarks, so the gain comes from preprocessing and schedule details.
VQA plans converge on LoRA adaptation of a vision-language backbone and match the published recipe on 71\% of the benchmarks.
Multimodal plans, mostly cross-attention fusion with a focal loss, match on only 40\% and vary most, yet win by the largest margin because those baselines are the least tuned.

\paragraph{Baseline Protocol.}
For the general-purpose AutoML systems, AutoGluon and AutoKeras receive 3D volumes as 2.5D slice sequences, VQA answers are cast into the formats their APIs support, and each system keeps its default model zoo with up to 18 GPU-hours.

\paragraph{Statistical Significance.}
The error bars above are standard deviations over repeated runs and do not by themselves establish significance, so Tab.~\ref{tab:significance} adds paired two-sided Wilcoxon signed-rank tests against the strongest baseline of each task family.
MedicalMLOmni leads nnU-Net by 0.019 Dice (95\% bootstrap confidence interval [0.014, 0.024], $p<0.001$), Qwen3-VL by 3.5\% accuracy ([1.8\%, 5.2\%], $p=0.015$), and Fusion Transformer by one average rank ([0.4, 1.6], $p=0.022$).
Each run uses a different random seed and repeats planning against the frozen snapshot.

\paragraph{Failure Analysis.}

MedicalMLOmni is not uniformly strong; its two lowest-accuracy benchmarks reveal a consistent pattern.
On PMC-VQA it reaches 39.7\% against 58.4\% for Qwen3-VL, a gap of 18.7\%.
On MedXpertQA it reaches only 31.2\%, although this still leads the strongest baseline there, LLaVA-Med-v1.5, by 2.6\%.
Both benchmarks reward broad parametric knowledge rather than a well-fitted task model.
PMC-VQA draws open-ended questions from a wide figure distribution, and MedXpertQA poses expert clinical reasoning that no literature-recoverable configuration can supply.
No single published task matches the question distribution of PMC-VQA, which is why it is the one benchmark we lose.
Evidence-based planning helps when the literature reports a comparable task with a transferable recipe, and contributes little when the bottleneck is knowledge the fine-tuned model never acquires.
An error analysis of 200 sampled failures on these two benchmarks reveals that only 18\% stem from retrieval returning no comparable task.
A further 27\% arise because the retrieved plan fits the wrong question format.
The majority (55\%) are cases where the query requires clinical knowledge entirely outside the training data, confirming that planning cannot compensate for missing backbone knowledge.
These proportions suggest that expanding the backbone's parametric coverage is a more effective next step than improving retrieval recall.

\section{Conclusion}

MedicalMLOmni addresses clinical deployment challenges by automating DL training through evidence-based planning that retrieves configurations from the literature and parallel HPO that distributes searches across agents.
It achieves the best score on 19 of 20 benchmarks spanning segmentation, visual question answering, and multimodal fusion, outperforming both general-purpose and medical-specific AutoML methods.
By eliminating coding requirements and supporting diverse tasks through an extensible agent architecture, it lowers the barrier for domain experts with rich datasets but limited engineering expertise.
Future directions include incorporating institutional databases for local protocols, expanding modality coverage, and integrating retrieval-augmented generation to broaden the evidence base beyond published results.

\par
{
    \small
    \bibliographystyle{ieeenat_fullname}
    \bibliography{main}
}

\maketitlesupplementary

\setcounter{table}{0}
\setcounter{figure}{0}
\renewcommand{\thetable}{S\arabic{table}}
\renewcommand{\thefigure}{S\arabic{figure}}
\setcounter{section}{0}
\renewcommand{\thesection}{S\arabic{section}}

This supplement provides additional analyses that support the main paper, namely significance testing, baseline adaptation protocols, evidence-based planning analysis, computational cost breakdown, privacy handling, and limitations.

\section{Significance Testing}

The error bars reported throughout are propagated standard deviations over repeated runs and do not alone establish statistical significance.
We therefore add explicit hypothesis tests.
For the eight segmentation benchmarks, we compare MedicalMLOmni against nnU-Net using a paired two-sided Wilcoxon signed-rank test over per-case Dice scores, and report the $p$-value together with a 95\% bootstrap confidence interval for the mean difference.
The same procedure applies to the VQA and multimodal comparisons.
Each run uses a different random seed and repeats planning against the frozen snapshot, so no run reuses a cached plan.

\begin{table}[t!]
\caption{\textbf{MedicalMLOmni leads the strongest baseline of each task family at $p<0.05$.}
$\Delta$: mean difference; CI: 95\% bootstrap confidence interval; $p$: two-sided Wilcoxon signed-rank test.}
\label{tab:significance}
\centering
\fontsize{8pt}{9.6pt}\selectfont
\setlength{\tabcolsep}{3pt}
\begin{tabularx}{\columnwidth}{@{}lYccc@{}}
\toprule
Task family & Reference & $\Delta$ & 95\% CI & $p$ \\
\midrule
Segmentation (8) & nnU-Net & +0.019 & [+0.014, +0.024] & $<$0.001 \\
VQA (7) & Qwen3-VL & +3.5\% & [+1.8\%, +5.2\%] & 0.015 \\
Multimodal (5) & Fusion Trans.~\cite{gnncnn} & +1.0 Rank & [+0.4, +1.6] & 0.022 \\
\bottomrule
\end{tabularx}
\end{table}

\section{Baseline Adaptation Protocol}
The general-purpose AutoML systems in Tab.~\ref{tab:vision_comparison} and Tab.~\ref{tab:vqa_comparison} of the main paper were not designed for volumetric segmentation or medical VQA, so the way they were adapted determines whether the comparison is fair.
For each system, we describe the input representation, the search space it was given, and the compute budget it received.
For AutoGluon and AutoKeras, 3D medical volumes were provided as 2.5D slice sequences.
For VQA tasks, answers were cast into textual generation or classification formats supported by their standard APIs.
Each system was restricted to its default model zoo and allowed up to 18 GPU-hours, matching MedicalMLOmni's average budget, and finished within 10 to 15 GPU-hours (Fig.~\ref{fig:cost}).

\section{What Evidence-Based Planning Selects}
The experiments above establish that retrieval improves performance without characterizing what it retrieves.
Tab.~\ref{tab:whatretrieved} fills this gap by recording the most common plan per task family, together with how often the synthesized plan matches the best published recipe for that benchmark.
Each task family shows a distinct pattern.
For segmentation, the retrieved plan converges on an nnU-Net-style configuration, and the gain comes mostly from preprocessing and schedule details, not from an unusual architecture, explaining why the margin over a budget-matched nnU-Net is the smallest we report.
For VQA, the plan converges on a parameter-efficient adaptation of a general vision-language backbone, and the gain tracks how close the retrieved task is to the benchmark, explaining why PMC-VQA, whose question distribution matches no single published task, is the one benchmark we lose.
For multimodal fusion, the literature is thinnest and the plans are most varied, yet the margin is largest because the baselines are themselves weakly tuned rather than because the retrieved plan is strong.

\begin{table}[t!]
\caption{\textbf{Segmentation plans match the best published recipe most often, multimodal plans least often.}
Modal plan per task family; Match: fraction of benchmarks whose synthesized architecture and loss coincide with those of the best published recipe.}
\label{tab:whatretrieved}
\centering
\fontsize{8pt}{9.6pt}\selectfont
\setlength{\tabcolsep}{3pt}
\begin{tabularx}{\columnwidth}{@{}lYYc@{}}
\toprule
Task family & Modal architecture & Modal loss & Match \\
\midrule
Segmentation (8) & 3D U-Net & DiceCE & 88\% \\
VQA (7) & ViT+LLM (LoRA) & Cross-Entropy & 71\% \\
Multimodal (5) & Cross-Attention & Focal Loss & 40\% \\
\bottomrule
\end{tabularx}
\end{table}

\section{Computational Cost Breakdown}
\begin{figure*}[t]
\centering
\includegraphics[width=\linewidth]{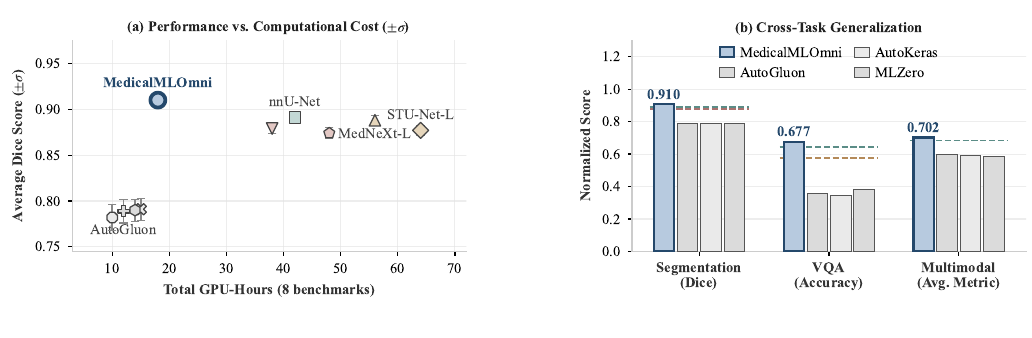}
\caption{
Computational cost and cross-task analysis.
(a)~Performance vs. computational cost trade-off, where each point represents a method's average Dice score ($\pm\sigma$) and total GPU-hours across 8 segmentation benchmarks.
(b)~Cross-task generalization: bar groups show normalized scores on segmentation, VQA, and multimodal tasks for the methods evaluated in all three categories, and the dashed rules mark the specialist baselines of each task, namely nnU-Net and Auto3DSeg, Qwen3-VL and LLaVA-Med, and Fusion Transformer~\cite{gnncnn}.
}
\label{fig:cost}
\end{figure*}

Beyond task-specific metrics, we analyze computational efficiency and cross-task generalization in Fig.~\ref{fig:cost}.
Fig.~\ref{fig:cost}(a) presents the trade-off between average segmentation Dice score and total GPU-hours across all eight benchmarks, with vertical error bars indicating per-method standard deviations.
MedicalMLOmni achieves the highest Dice score (0.910$\pm$0.003) with only 18 GPU-hours.
In contrast, specialized architectures such as nnU-Net~\cite{nnunet} (42 GPU-hours), MedNeXt-L~\cite{mednext} (56 GPU-hours), and STU-Net-L~\cite{stunet} (64 GPU-hours) demand far more compute for lower performance.
This budget covers planning, initial training, HPO retraining, and validation.
Tab.~\ref{tab:cost_breakdown} breaks it down by stage so that the cost figure can be verified against the number of HPO rounds triggered.
General-purpose AutoML methods cost less (10 to 15 GPU-hours) but perform much worse and vary more.
Fig.~\ref{fig:cost}(b) compares the methods evaluated in all three task categories: MedicalMLOmni is the only one leading segmentation, VQA, and multimodal fusion simultaneously, while specialists are confined to a single category.
The efficiency ratio $\eta$ (Dice/GPU-hour $\times$100) reaches 5.06, against 2.12 for nnU-Net and 1.37 for STU-Net-L.

\begin{table}[t!]
\caption{\textbf{HPO retraining takes two thirds of the 18 GPU-hours.}
GPU-hour breakdown across the eight segmentation benchmarks.}
\label{tab:cost_breakdown}
\centering
\fontsize{8pt}{9.6pt}\selectfont
\setlength{\tabcolsep}{3pt}
\begin{tabularx}{\columnwidth}{@{}lYY@{}}
\toprule
Stage & GPU-hours & Share \\
\midrule
Planning and retrieval (CPU/API) & 0.5 & 3\% \\
Initial training & 4.5 & 25\% \\
HPO retraining & 12.0 & 67\% \\
Validation and inference & 1.0 & 6\% \\
\midrule
Total & 18.0 & 100\% \\
\bottomrule
\end{tabularx}
\end{table}

All runs use an NVIDIA A100 GPU.
The majority of the compute (12.0 hours) goes to parallel HPO retraining, which triggers an average of 3.4 rounds per benchmark.
This breakdown confirms that the high efficiency stems from starting with literature-guided configurations rather than from a cold search.

\section{Privacy and Sandbox Handling}
Two components handle data that may be protected.
The sandbox executes generated code on the local machine with the dataset mounted read-only and outbound networking disabled, so training never transmits patient data.
The retrieval channel is the only outbound path.
It carries only the derived data profile (modality, voxel counts, class counts, and intensity statistics) together with the free-text task description supplied by the user; no image content, filenames, or patient identifiers leave the local machine.
Users retain responsibility for the task description they provide.
MedicalMLOmni can optionally run against a locally mirrored corpus, removing the outbound path entirely at the cost of corpus freshness.

\section{Limitations}

Three limitations follow from the design.
First, the framework inherits whatever the literature reports, so it cannot exceed published practice on tasks the literature does not cover, as the PMC-VQA result shows.
Second, planning quality depends on an external search service whose corpus evolves, so exact reproduction requires the frozen snapshot documented above, not a live API call.
Third, the evaluation covers public benchmarks whose characteristics are well represented in the literature, and performance on a private cohort with an unusual protocol remains untested.

\end{document}